\documentclass{article}

\usepackage{colacl,psfig,piclatex,rotate}

\title{TnT --- A Statistical Part-of-Speech Tagger}

\author{Thorsten Brants\\
        Saarland University\\ Computational Linguistics\\
        D-66041 Saarbr\"ucken, Germany\\
        {\tt thorsten@coli.uni-sb.de}\\[1ex]
        In {\em Proceedings of the Sixth Applied Natural Language
        Processing Conference ANLP-2000},\\ 
        April 29 -- May 3, 2000, Seattle, WA.}

\def\setgray#1{
\ignorespaces}
\def\unsetgray{
\ignorespaces}
\def\argmax{\mathop{\rm argmax}}
\def\tnt{{TnT}}
\expandafter\ifx\csname rotate\endcsname\relax
\makeatletter
\newbox\@rotbox
\def\rotate{\@ifnextchar[{\@rotate}{\@rotate[l]}}
\def\@rotate[#1]#2{\setbox\@rotbox=\hbox{#2}\@nameuse{@rot#1}\@rotbox}
\makeatother
\fi

\begin{document}

\maketitle

\begin{abstract}
  Trigrams'n'Tags (\tnt) is an efficient statistical part-of-speech
  tagger. Contrary to claims found elsewhere in the literature, we
  argue that a tagger based on Markov models performs at least as well
  as other current approaches, including the Maximum Entropy
  framework.  A recent comparison has even shown that
  \tnt\ performs significantly better for the tested corpora. We
  describe the basic model of \tnt, the techniques used for
  smoothing and for handling unknown words. Furthermore, we present
  evaluations on two corpora.
\end{abstract}

\section{Introduction}
\label{sec:intro}

A large number of current language processing systems use a
part-of-speech tagger for pre-processing. The tagger assigns a (unique
or ambiguous) part-of-speech tag to each token in the input and passes
its output to the next processing level, usually a parser. Furthermore,
there is a large interest in part-of-speech tagging for corpus
annotation projects, who create valuable linguistic resources by
a combination of automatic processing and human correction.

For both applications, a tagger with the highest possible accuracy is
required. The debate about which paradigm solves the part-of-speech
tagging problem best is not finished. Recent comparisons of approaches
that can be trained on corpora \cite{Halteren:ea:98,Volk:Schneider:98}
have shown that in most cases statistical aproaches
\cite{Cutting:ea:92,Schmid:95,Ratnaparkhi:96} yield better results than
finite-state, rule-based, or memory-based taggers
\cite{Brill:93diss,Daelemans:ea:96}. They are only surpassed by
combinations of different systems, forming a ``voting tagger''.

Among the statistical approaches, the Maximum Entropy framework has a
very strong position. Nevertheless, a recent independent comparison of
7 taggers \cite{Zavrel:Daelemans:99} has shown that another approach
even works better: Markov models combined with a good smoothing
technique and with handling of unknown words. This tagger, \tnt, not
only yielded the highest accuracy, it also was the fastest both in
training and tagging.

The tagger comparison was organized as a ``black-box test'': set the
same task to every tagger and compare the outcomes. This paper
describes the models and techniques used by \tnt\ together with the
implementation.

The reader will be surprised how simple the underlying model is.  The
result of the tagger comparison seems to support the maxime ``the
simplest is the best''. However, in this paper we clarify a number of
details that are omitted in major previous publications concerning
tagging with Markov models. As two examples, \cite{Rabiner:89} and
\cite{Charniak:ea:93} give good overviews of the techniques and
equations used for Markov models and part-of-speech tagging, but they
are not very explicit in the details that are needed for their
application. We argue that it is not only the choice of the general
model that determines the result of the tagger but also the various
``small'' decisions on alternatives.

The aim of this paper is to give a detailed account of the techniques
used in \tnt. Additionally, we present results of the
tagger on the NEGRA corpus \cite{Brants:ea:99} and the Penn Treebank
\cite{Marcus:ea:93}.  The Penn Treebank results reported here for the
Markov model approach are at least equivalent to those reported for
the Maximum Entropy approach in \cite{Ratnaparkhi:96}.  For a
comparison to other taggers, the reader is referred to
\cite{Zavrel:Daelemans:99}.

\section{Architecture}
\label{sec:archi}

\subsection{The Underlying Model}

\tnt\ uses second order Markov models for part-of-speech
tagging. The states of the model represent tags, outputs
represent the words. Transition probabilities depend on the states,
thus pairs of tags. Output probabilities only depend on the most
recent category. To be explicit, we calculate
\begin{equation}
\argmax\limits_{t_1\dots t_T} 
  \left[\prod_{i=1}^{T}P(t_i|t_{i-1},t_{i-2})P(w_i|t_i)\right] P(t_{T+1}|t_T)
\end{equation}
for a given sequence of words $w_1\dots w_T$ of length $T$.
$t_1\dots t_T$ are elements of the tagset, the additional tags
$t_{-1}$, $t_{0}$, and $t_{T+1}$ are beginning-of-sequence and
end-of-sequence markers.  Using these additional tags, even if they
stem from rudimentary processing of punctuation marks, slightly
improves tagging results. This is different from formulas presented in
other publications, which just stop with a ``loose end'' at the last
word. If sentence boundaries are not marked in the input, \tnt\ adds
these tags if it encounters one of [.!?;] as a token.

Transition and output probabilities are estimated from a tagged
corpus. As a first step, we use the maximum likelihood probabilities
$\hat P$ which are derived from the relative frequencies:
\begin{eqnarray}
\mbox{Unigrams:} && \hat P(t_3) = \frac{f(t_3)}{N} \\
\label{eq:uni}
\mbox{Bigrams:}  && \hat P(t_3|t_2) = \frac{f(t_2,t_3)}{f(t_2)} \\
\label{eq:bi}
\mbox{Trigrams:} && \hat P(t_3|t_1,t_2) = \frac{f(t_1,t_2,t_3)}{f(t_1,t_2)} \\
\label{eq:tri}
\mbox{Lexical:}  && \hat P(w_3|t_3) = \frac{f(w_3,t_3)}{f(t_3)}
\end{eqnarray}
for all $t_1$, $t_2$, $t_3$ in the tagset and $w_3$ in the lexicon.
$N$ is the total number of tokens in the training corpus.  We define a
maximum likelihood probability to be zero if the corresponding
nominators and denominators are zero. As a second step, contextual
frequencies are smoothed and lexical frequences are completed by
handling words that are not in the lexicon (see below).

\subsection{Smoothing}

Trigram probabilities generated from a corpus usually cannot
directly be used because of the sparse-data problem. This means that
there are not enough instances for each trigram to reliably estimate
the probability. Furthermore, setting a probability to zero because
the corresponding trigram never occured in the corpus has an undesired
effect. It causes the probability of a complete sequence to be set to
zero if its use is necessary for a new text sequence, thus makes it
impossible to rank different sequences containing a zero probability.

The smoothing paradigm that delivers the best results in \tnt\ is linear
interpolation of unigrams, bigrams, and trigrams. Therefore, we
estimate a trigram probability as follows:
\begin{equation}
  P(t_3|t_1,t_2) = \lambda_1 \hat P(t_3) + \lambda_2 \hat P(t_3|t_2) +
  \lambda_3 \hat P(t_3|t_1,t_2)
\end{equation}
$\hat P$ are maximum likelihood estimates of the probabilities, and 
$\lambda_1 + \lambda_2 + \lambda_3 = 1$, so $P$ again represent 
probability distributions.

We use the context-independent variant of linear interpolation, i.e.,
the values of the $\lambda$s do not depend on the particular trigram.
Contrary to intuition, this yields better results than the
context-dependent variant. Due to sparse-data problems, one cannot
estimate a different set of $\lambda$s for each trigram. Therefore, it
is common practice to group trigrams by frequency and estimate tied
sets of $\lambda$s. However, we are not aware of any publication that
has investigated frequency groupings for linear interpolation in
part-of-speech tagging. All groupings that we have tested yielded at
most equivalent results to context-independent linear interpolation.
Some groupings even yielded worse results. The tested groupings included a)
one set of $\lambda$s for each frequency value and b) two classes (low
and high frequency) on the two
ends of the scale, as well as several groupings in between and several
settings for partitioning the classes.

The values of $\lambda_1$, $\lambda_2$, and $\lambda_3$ are estimated
by deleted interpolation. This technique successively removes each
trigram from the training corpus and estimates best values for the
$\lambda$s from all other $n$-grams in the corpus. Given the frequency
counts for uni-, bi-, and trigrams, the weights can be very
efficiently determined with a processing time linear in the number of
different trigrams. The algorithm is given in figure
\ref{fig:lambdaalg}. Note that subtracting 1 means taking unseen data
into account. Without this subtraction the model would overfit the
training data and would generally yield worse results.

\begin{figure*}
\hrule
\medskip
\hspace*{2cm}\begin{minipage}{12cm}
\begin{tt}
\begin{tabbing}
\ \ \ \ \=\ \ \ \ \=\ \ \ \ \=\kill
set $\lambda_1 = \lambda_2 = \lambda_3 = 0$\\[.5ex]
foreach trigram $t_1,t_2,t_3$ with $f(t_1,t_2,t_3)>0$\\[.5ex]
\>  depending on the maximum of the following three values:\\[.5ex]
\>  \>  case {\large$\frac{f(t_1,t_2,t_3)-1}{f(t_1,t_2)-1}$}: increment $\lambda_3$ by $f(t_1,t_2,t_3)$\\[1ex]
\>  \>  case {\large$\frac{f(t_2,t_3)-1}{f(t_2)-1}$}: increment $\lambda_2$ by $f(t_1,t_2,t_3)$ \\[1ex]
\>  \>  case {\large$\frac{f(t_3)-1}{N-1}$}: increment $\lambda_1$ by $f(t_1,t_2,t_3)$\\[.5ex]
\>  end\\
end\\
normalize $\lambda_1,\lambda_2,\lambda_3$
\end{tabbing}
\end{tt}
\end{minipage}
\medskip
\hrule
\caption{Algorithm for calculting the weights for context-independent
  linear interpolation $\lambda_1,\lambda_2,\lambda_3$ when the
  $n$-gram frequencies are known. $N$ is the size of the corpus. If
  the denominator in one of the expressions is 0, we define the result
  of that expression to be 0.}
\label{fig:lambdaalg}
\end{figure*}

\subsection{Handling of Unknown Words}

Currently, the method of handling unknown words that seems to work
best for inflected languages is a suffix analysis as proposed in
\cite{Samuelsson:93}.  Tag probabilities are set according to the
word's ending. The suffix is a strong predictor for word classes,
e.g., words in the Wall Street Journal part of the Penn Treebank
ending in {\em able} are adjectives ({\sf JJ}) in 98\% of the cases
(e.g.\ fashionable, variable) , the rest of 2\% are nouns (e.g.\ 
cable, variable).

The probability distribution for a
particular suffix is generated from all words in the
training set that share the same suffix of some predefined maximum
length. The term suffix as used here means ``final sequence of
characters of a word'' which is not necessarily a linguistically
meaningful suffix.

Probabilities are smoothed by successive abstraction.  This
calculates the probability of a tag $t$ given the last $m$ letters
$l_i$ of an $n$ letter word: $P(t|l_{n-m+1}, \dots l_n)$. The sequence
of increasingly more general contexts omits more and more characters
of the suffix, such that $P(t|l_{n-m+2}, \dots, l_n)$, $P(t|l_{n-m+3},
\dots, l_n)$, \dots, $P(t)$ are used for smoothing.  The recursion
formula is
\[
P(t|l_{n-i+1}, \dots l_n)  \hspace*{5cm}
\]
\begin{equation}
 = \frac{\hat P(t|l_{n-i+1}, \dots l_n) +
  \theta_i P(t|l_{n-i}, \dots, l_n)}{1+\theta_i}
\end{equation}
for $i = m \dots 0$,
using the maximum likelihood estimates $\hat P$ from frequencies
in the lexicon, weights $\theta_i$ and the initialization
\begin{equation}
  P(t) = \hat P(t).
\end{equation}
The maximum likelihood estimate for a suffix of length $i$ is derived
from corpus frequencies by
\begin{equation}
\hat P(t|l_{n-i+1}, \dots l_n) = 
    \frac{f(t, l_{n-i+1}, \dots l_n)}{f(l_{n-i+1}, \dots l_n)}
\end{equation}
For the Markov model, we need the inverse conditional probabilities
$P(l_{n-i+1}, \dots l_n|t)$ which are obtained by Bayesian inversion.

A theoretical motivated argumentation uses the standard deviation of the
maximum likelihood probabilities for the weights $\theta_i$
\cite{Samuelsson:93}. 

This leaves room for interpretation. 

1) One has to identify a good
value for $m$, the longest suffix used.
The approach taken for \tnt\ is the following: $m$ depends
on the word in question. We use the longest suffix that we can find in
the training set (i.e., for which the frequency is greater than or equal to
1), but at most 10 characters. This is an empirically determined choice.

2) We use a context-independent approach for $\theta_i$, as we did for
the contextual weights $\lambda_i$. It turned out to be a good choice
to set all $\theta_i$ to the standard deviation of the unconditioned
maximum likelihood probabilities of the tags in the training corpus,
i.e., we set
\begin{equation}
\theta_i = \frac{1}{s-1}\sum_{j=1}^{s}(\hat P(t_j) - \bar P)^2
\end{equation}
for all $i=0\dots m-1$, using a tagset of $s$ tags and the average
\begin{equation}
\bar P = \frac{1}{s} \sum_{j=1}^{s} \hat P(t_j)
\end{equation}
This usually yields values in the range 0.03 \dots 0.10.

3) We use different estimates for uppercase and lowercase words,
i.e., we maintain two different suffix tries depending on the
capitalization of the word. This information improves the tagging results.

4) Another freedom concerns the choice of the words in the lexicon that
should be used for suffix handling. Should we use all words, or are
some of them better suited than others? Accepting that unknown words
are most probably infrequent, one can argue that using suffixes
of infrequent words in the lexicon is a better approximation for
unknown words than using suffixes of frequent words. Therefore, we
restrict the procedure of suffix handling to words with a frequency
smaller than or equal to some threshold value. Empirically, 10 turned
out to be a good choice for this threshold.

\subsection{Capitalization}
\label{sec:capitalization}

Additional information that turned out to be useful for the
disambiguation process for several corpora and tagsets is
capitalization information. Tags are usually not informative about
capitalization, but probability distributions of tags around
capitalized words are different from those not capitalized. The effect
is larger for English, which only capitalizes proper names, and smaller
for German, which capitalizes all nouns.

We use flags $c_i$ that are true if $w_i$ is a capitalized word and
false otherwise. These flags are added to the contextual probability
distributions. Instead of
\begin{equation}
  P(t_3|t_1,t_2)
\end{equation}
we use
\begin{equation}
  P(t_3,c_3|t_1,c_1,t_2,c_2)
\end{equation}
and equations (\ref{eq:uni}) to (\ref{eq:tri}) are updated
accordingly. This is equivalent to doubling the size of the tagset and
using different tags depending on capitalization.

\subsection{Beam Search}

The processing time of the Viterbi algorithm \cite{Rabiner:89} can be
reduced by introducing a beam search. Each state that receives a
$\delta$ value smaller than the largest $\delta$ divided by some
threshold value $\theta$ is excluded from further processing.  While
the Viterbi algorithm is guaranteed to find the sequence of states
with the highest probability, this is no longer true when beam search
is added. Nevertheless, for practical purposes and the right choice of
$\theta$, there is virtually no difference between the algorithm with
and without a beam. Empirically, a value of $\theta=1000$ turned out
to approximately double the speed of the tagger without affecting the
accuracy. 

The tagger currently tags between 30,000 and 60,000 tokens per second
(including file I/O) on a Pentium 500 running Linux. The speed mainly
depends on the percentage of unknown words and on the average ambiguity
rate.

\section{Evaluation}

We evaluate the tagger's performance under several aspects. First of
all, we determine the tagging accuracy averaged over ten
iterations. The overall accuracy, as well as separate accuracies for
known and unknown words are measured. 

Second, learning curves are
presented, that indicate the performance when using training corpora
of different sizes, starting with as few as 1,000 tokens and ranging
to the size of the entire corpus (minus the test set).

An important characteristic of statistical taggers is that they not
only assign tags to words but also probabilities in order to rank
different assignments. We distinguish reliable from unreliable
assignments by the quotient of the best and second best
assignments\footnote{By definition, this quotient is $\infty$ if there
  is only one possible tag for a given word.}. All assignments for
which this quotient is larger than some threshold are regarded as
reliable, the others as unreliable. As we will see below, accuracies
for reliable assignments are much higher.

The tests are performed on partitions of the corpora that use 90\% as
training set and 10\% as test set, so that the test data is guaranteed
to be unseen during training. Each result is obtained by repeating the
experiment 10 times with different partitions and averaging the single
outcomes. 

In all experiments, contiguous test sets are used. The alternative is
a round-robin procedure that puts every 10th sentence into the test
set. We argue that contiguous test sets yield more realistic results
because completely unseen articles are tagged. Using the round-robin
procedure, parts of an article are already seen, which significantly
reduces the percentage of unknown words. Therefore, we expect even
higher results when testing on every 10th sentence instead of
a contiguous set of 10\%.

In the following, accuracy denotes the number of correctly assigned
tags divided by the number of tokens in the corpus processed. The
tagger is allowed to assign exactly one tag to each token.

We distinguish the overall accuracy, taking into account all tokens in
the test corpus, and separate accuracies for known and unknown tokens.
The latter are interesting, since usually unknown tokens are much more
difficult to process than known tokens, for which a list of valid tags
can be found in the lexicon.

\subsection{Tagging the NEGRA corpus}

The German NEGRA corpus consists of 20,000 sentences (355,000 tokens)
of newspaper texts (Frankfurter Rundschau) that are annotated with
parts-of-speech and predicate-argument structures \cite{Skut:ea:97a}.
It was developed at the Saarland University in
Saarbr\"ucken\footnote{For availability, please check \\{\tt
    http://www.coli.uni-sb.de/sfb378/negra-corpus}}. Part of it was
tagged at the IMS Stuttgart. This evaluation only uses the
part-of-speech annotation and ignores structural annotations.

Tagging accuracies for the NEGRA corpus are shown in table
\ref{tab:negraacc}. 

Figure \ref{fig:negra-learn} shows the learning curve of the tagger,
i.e., the accuracy depending on the amount of training data. Training
length is the number of tokens used for training. Each training length
was tested ten times, training and test sets were randomly chosen and
disjoint, results were averaged. The training length is given on a
logarithmic scale. 

It is remarkable that tagging accuracy for known words is very high
even for very small training corpora. This means that we have a good
chance of getting the right tag if a word is seen at least once during
training. 
Average percentages of unknown tokens are shown in the bottom line of
each diagram. 

\begin{figure*}
\let\fnamebak\figurename
\let\figurename\tablename
\caption{Part-of-speech tagging accuracy for the NEGRA corpus,
  averaged over 10 test runs, training and test set are disjoint.
  The table shows the percentage of unknown tokens, separate
  accuracies and standard deviations for known and unknown tokens, as
  well as the overall accuracy.}
\label{tab:negraacc}
\smallskip
\hrule
\begin{center}
\begin{tabular}{l|r|rr|rr|rr}
             & percentage & \multicolumn{2}{c|}{known} & \multicolumn{2}{c|}{unknown} & \multicolumn{2}{c}{overall} \\
             & unknowns   & acc.   & $\sigma$ & acc.   & $\sigma$ & acc.   & $\sigma$ \\
\hline
NEGRA corpus & 11.9\%     & 97.7\% & 0.23     & 89.0\% & 0.72     & 96.7\% & 0.29 \\
\end{tabular}
\end{center}
\let\figurename\fnamebak
\hrule
\bigskip
\hrule
\bigskip
\begin{center}
{\bf NEGRA Corpus: POS Learning Curve}

\bigskip
\hspace*{0pt}
\beginpicture
\setcoordinatesystem units <30mm,.8mm>
\setplotarea x from 3 to 6, y from 50 to 100
\setbox0=\hbox{\raise .4ex\hbox{\scriptsize 320}}
\axis bottom ticks withvalues 1 2 5 10 20 50 100 200 \usebox0 500 1000 / 
                           at 3 3.3 3.7 4 4.3 4.7 5 5.3 5.505 5.7 6 / /
\axis left ticks numbered from 50 to 100 by 10 /
\put {\small 50.8} [t] <0mm,-8.5mm> at 3 50
\put {\small 46.4} [t] <0mm,-8.5mm> at 3.3 50 
\put {\small 41.4} [t] <0mm,-8.5mm> at 3.7 50
\put {\small 36.0} [t] <0mm,-8.5mm> at 4 50
\put {\small 30.7} [t] <0mm,-8.5mm> at 4.3 50
\put {\small 23.0} [t] <0mm,-8.5mm> at 4.7 50
\put {\small 18.3} [t] <0mm,-8.5mm> at 5 50
\put {\small 14.3} [t] <0mm,-8.5mm> at 5.3 50
\put {\tiny 11.9} [t] <0mm,-8.5mm> at 5.505 50
\put {\setgray{0.5}\small 10.3\unsetgray} [t] <0mm,-8.5mm> at 5.7 50
\put {\setgray{0.5}\small 8.4\unsetgray} [t] <0mm,-8.5mm> at 6 50
\put {\small avg.\ percentage unknown} [tl] <3mm,-8.5mm> at 6.05 50
\linethickness=.4pt
\putbar breadth <0pt> from 3 60 to 6 60
\putbar breadth <0pt> from 3 70 to 6 70
\putbar breadth <0pt> from 3 80 to 6 80
\putbar breadth <0pt> from 3 90 to 6 90
\putbar breadth <0pt> from 3 100 to 6 100
\put {\rotate[l]{Accuracy}} [r] <-9mm,0mm> at 3 75
\put {$\times1000$ Training Length} [lt] <4mm,-3mm> at 6 50
\put {\rule{10mm}{1.6pt}} [l] <5mm,0mm> at 6 94
\put {Overall} [l] <18mm,0mm> at 6 94
\put {\small\tabcolsep=0pt\begin{tabular}{lcr}min&=&78.1\%\\
                     max&=&96.7\%\\\end{tabular}} [l] <18mm,0mm> at 6 86
\put {\rule{10mm}{.8pt}} [l] <5mm,0mm> at 6 79
\put {$\bullet$} [l] <9mm,0mm> at 6 79
\put {Known} [l] <18mm,0mm> at 6 79
\put {\small\tabcolsep=0pt\begin{tabular}{lcr}min&=&95.7\%\\
                     max&=&97.7\%\\\end{tabular}} [l] <18mm,0mm> at 6 70
\put {\rule{10mm}{.8pt}} [l] <5mm,0mm> at 6 63
\put {$\circ$} [l] <9mm,0mm> at 6 63
\put {Unknown} [l] <18mm,0mm> at 6 63
\put {\small\tabcolsep=0pt\begin{tabular}{lcr}min&=&61.2\%\\
                     max&=&89.0\%\\\end{tabular}} [l] <18mm,0mm> at 6 54
\setlinear
\setplotsymbol({\rule{1.6pt}{1.6pt}})
\plot
        3.000   78.12   %% 1000
        3.301   83.36   %% 2000
        3.699   88.06   %% 5000
        4.000   90.54   %% 10000
        4.301   92.61   %% 20000
        4.699   94.42   %% 50000
        5.000   95.58   %% 100000
        5.301   96.23   %% 200000
        5.505   96.65   %% 330000
/
\setplotsymbol({\rule{.8pt}{.8pt}})
\plot
        3.000   95.68   %% 1000
        3.301   96.14   %% 2000
        3.699   96.34   %% 5000
        4.000   96.68   %% 10000
        4.301   97.01   %% 20000
        4.699   97.24   %% 50000
        5.000   97.46   %% 100000
        5.301   97.57   %% 200000
        5.505   97.68   %% 330000
/
\multiput {$\bullet$} at
        3.000   95.68   %% 1000
        3.301   96.14   %% 2000
        3.699   96.34   %% 5000
        4.000   96.68   %% 10000
        4.301   97.01   %% 20000
        4.699   97.24   %% 50000
        5.000   97.46   %% 100000
        5.301   97.57   %% 200000
        5.505   97.68   %% 330000
/
\plot
        3.000   61.24   %% 1000
        3.301   68.52   %% 2000
        3.699   75.60   %% 5000
        4.000   78.90   %% 10000
        4.301   81.87   %% 20000
        4.699   84.53   %% 50000
        5.000   86.93   %% 100000
        5.301   88.24   %% 200000
        5.505   89.02   %% 330000
/
\multiput {$\circ$} at
        3.000   61.24   %% 1000
        3.301   68.52   %% 2000
        3.699   75.60   %% 5000
        4.000   78.90   %% 10000
        4.301   81.87   %% 20000
        4.699   84.53   %% 50000
        5.000   86.93   %% 100000
        5.301   88.24   %% 200000
        5.505   89.02   %% 330000
/
\endpicture
\end{center}
\hrule
\caption{Learning curve for tagging the NEGRA corpus. The training sets
  of variable sizes as well as test sets of 30,000 tokens were
  randomly chosen. Training and test sets were disjoint, the procedure
  was repeated 10 times and results were averaged. Percentages of
  unknowns for 500k and 1000k training are determined from an untagged
  extension.}
\label{fig:negra-learn}
\bigskip
\hrule
\bigskip
\begin{center}
{\bf NEGRA Corpus: Accuracy of reliable assignments}

\bigskip
\hspace*{0pt}
\beginpicture
\setcoordinatesystem units <23mm,10mm>
\setplotarea x from 0 to 4, y from 96 to 100
\axis bottom ticks withvalues 1 2 5 10 20 50 100 500 2000 10000 / 
                           at 0 .3 .7 1 1.3 1.7 2 2.7 3.3 4 / /
\axis bottom ticks at 2.3 3 3.7 / /
\axis left ticks numbered from 96 to 100 by 1 /
\linethickness=.4pt
\putbar breadth <0pt> from 0 97 to 4 97
\putbar breadth <0pt> from 0 98 to 4 98
\putbar breadth <0pt> from 0 99 to 4 99
\putbar breadth <0pt> from 0 100 to 4 100
\put {\rotate[l]{Accuracy}} [r] <-9mm,0mm> at 0 98
\put {threshold $\theta$} [lt] <7mm,-2.6mm> at 4 96
\put {\small 100} [t] <0mm,-8.5mm> at 0 96
\put {\small 97.9} [t] <0mm,-8.5mm> at .3 96 
\put {\small 95.1} [t] <0mm,-8.5mm> at .7 96 
\put {\small 92.7} [t] <0mm,-8.5mm> at 1 96 
\put {\small 90.3} [t] <0mm,-8.5mm> at 1.3 96 
\put {\small 86.8} [t] <0mm,-8.5mm> at 1.7 96 
\put {\small 84.1} [t] <0mm,-8.5mm> at 2 96 
\put {\small 81.0} [t] <0mm,-8.5mm> at 2.3 96 
\put {\small 76.1} [t] <0mm,-8.5mm> at 2.7 96 
\put {\small 71.9} [t] <0mm,-8.5mm> at 3 96 
\put {\small 68.3} [t] <0mm,-8.5mm> at 3.3 96 
\put {\small 64.1} [t] <0mm,-8.5mm> at 3.7 96 
\put {\small 62.0} [t] <0mm,-8.5mm> at 4 96 
\put {\% cases reliable} [lt] <7mm,-8.5mm> at 4 96
\put {\small --} [t] <0mm,-13.5mm> at 0 96
\put {\small 53.5} [t] <0mm,-13.5mm> at .3 96 
\put {\small 62.9} [t] <0mm,-13.5mm> at .7 96 
\put {\small 69.6} [t] <0mm,-13.5mm> at 1 96 
\put {\small 74.5} [t] <0mm,-13.5mm> at 1.3 96 
\put {\small 79.8} [t] <0mm,-13.5mm> at 1.7 96 
\put {\small 82.7} [t] <0mm,-13.5mm> at 2 96 
\put {\small 85.2} [t] <0mm,-13.5mm> at 2.3 96 
\put {\small 88.0} [t] <0mm,-13.5mm> at 2.7 96 
\put {\small 89.6} [t] <0mm,-13.5mm> at 3 96 
\put {\small 90.8} [t] <0mm,-13.5mm> at 3.3 96 
\put {\small 91.8} [t] <0mm,-13.5mm> at 3.7 96 
\put {\small 92.2} [t] <0mm,-13.5mm> at 4 96 
\put {acc. of complement} [lt] <7mm,-13mm> at 4 96
\put {\tabcolsep=0pt\begin{tabular}{llcr}
    \hbox to 0pt{\hspace*{4mm}$\diamond$\hss}%
    \rule[.5ex]{10mm}{1.6pt}\ \ \ & \multicolumn{3}{l}{Reliable} \\
                             & \small min &\small=&\small 96.7\% \\
                             & \small max &\small=&\small 99.4\% \\
    \end{tabular}} [lb] <5mm,0mm> at 4 98
\setlinear
\setplotsymbol({\rule{1.6pt}{1.6pt}})
\plot
        0.000   96.74   %% 1
        0.301   97.61   %% 2
        0.699   98.42   %% 5
        1.000   98.81   %% 10
        1.301   99.07   %% 20
        1.699   99.25   %% 50
        2.000   99.33   %% 100
        2.301   99.38   %% 200
        2.699   99.42   %% 500
        3.000   99.43   %% 1000
        3.301   99.43   %% 2000
        3.699   99.42   %% 5000
        4.000   99.41   %% 10000
/
\multiput {$\diamond$} at
        0.000   96.74   %% 1
        0.301   97.61   %% 2
        0.699   98.42   %% 5
        1.000   98.81   %% 10
        1.301   99.07   %% 20
        1.699   99.25   %% 50
        2.000   99.33   %% 100
        2.301   99.38   %% 200
        2.699   99.42   %% 500
        3.000   99.43   %% 1000
        3.301   99.43   %% 2000
        3.699   99.42   %% 5000
        4.000   99.41   %% 10000
/
\endpicture
\end{center}
\hrule
\caption{Tagging accuracy for the NEGRA corpus when separating
  reliable and unreliable assignments. The curve shows accuracies for
  reliable assignments. The numbers at the bottom line
  indicate the percentage of reliable assignments and the accuracy of the
  complement set (i.e., unreliable assignments).}
\label{fig:negra-cond}
\end{figure*}

\begin{figure*}
\let\fnamebak\figurename
\let\figurename\tablename
\caption{Part-of-speech tagging accuracy for the Penn Treebank.
  The table shows the percentage of unknown tokens, separate
  accuracies and standard deviations for known and unknown tokens, as
  well as the overall accuracy.}
\label{tab:pennacc}
\smallskip
\hrule
\begin{center}
\begin{tabular}{l|r|rr|rr|rr}
             & percentage & \multicolumn{2}{c|}{known} & \multicolumn{2}{c|}{unknown} & \multicolumn{2}{c}{overall} \\
             & unknowns   & acc.   & $\sigma$ & acc.   & $\sigma$ & acc.   & $\sigma$ \\
\hline
Penn Treebank&  2.9\%     & 97.0\% & 0.15     & 85.5\% & 0.69     & 96.7\% & 0.15 \\
\end{tabular}
\end{center}
\hrule
\let\figurename\fnamebak
\bigskip
\hrule
\bigskip
\begin{center}
{\bf Penn Treebank: POS Learning Curve}

\bigskip
\hspace*{0pt}
\beginpicture
\setcoordinatesystem units <30mm,.8mm>
\setplotarea x from 3 to 6, y from 50 to 100
\axis bottom ticks withvalues 1 2 5 10 20 50 100 200 500 1000 / 
                           at 3 3.3 3.7 4 4.3 4.7 5 5.3 5.7 6 / /
\axis left ticks numbered from 50 to 100 by 10 /
\put {\small 50.3} [t] <0mm,-8.5mm> at 3 50
\put {\small 42.8} [t] <0mm,-8.5mm> at 3.3 50 
\put {\small 33.4} [t] <0mm,-8.5mm> at 3.7 50
\put {\small 26.8} [t] <0mm,-8.5mm> at 4 50
\put {\small 20.2} [t] <0mm,-8.5mm> at 4.3 50
\put {\small 13.2} [t] <0mm,-8.5mm> at 4.7 50
\put {\small 9.8} [t] <0mm,-8.5mm> at 5 50
\put {\small 7.0} [t] <0mm,-8.5mm> at 5.3 50
\put {\small 4.4} [t] <0mm,-8.5mm> at 5.7 50
\put {\small 2.9} [t] <0mm,-8.5mm> at 6 50
\put {\small avg.\ percentage unknown} [tl] <3mm,-8.5mm> at 6.05 50
\linethickness=.4pt
\putbar breadth <0pt> from 3 60 to 6 60
\putbar breadth <0pt> from 3 70 to 6 70
\putbar breadth <0pt> from 3 80 to 6 80
\putbar breadth <0pt> from 3 90 to 6 90
\putbar breadth <0pt> from 3 100 to 6 100
\put {\rotate[l]{Accuracy}} [r] <-9mm,0mm> at 3 75
\put {$\times1000$ Training Length} [lt] <4mm,-3mm> at 6 50
\put {\rule{10mm}{1.6pt}} [l] <5mm,0mm> at 6 94
\put {Overall} [l] <18mm,0mm> at 6 95
\put {\small\tabcolsep=0pt\begin{tabular}{lcr}min&=&78.6\%\\
                     max&=&96.7\%\\\end{tabular}} [l] <18mm,0mm> at 6 87
\put {\rule{10mm}{.8pt}} [l] <5mm,0mm> at 6 79
\put {$\bullet$} [l] <9mm,0mm> at 6 79
\put {Known} [l] <18mm,0mm> at 6 79
\put {\small\tabcolsep=0pt\begin{tabular}{lcr}min&=&95.2\%\\
                     max&=&97.0\%\\\end{tabular}} [l] <18mm,0mm> at 6 71
\put {\rule{10mm}{.8pt}} [l] <5mm,0mm> at 6 63
\put {$\circ$} [l] <9mm,0mm> at 6 63
\put {Unknown} [l] <18mm,0mm> at 6 63
\put {\small\tabcolsep=0pt\begin{tabular}{lcr}min&=&62.2\%\\
                     max&=&85.5\%\\\end{tabular}} [l] <18mm,0mm> at 6 55
\setlinear
\setplotsymbol({\rule{1.6pt}{1.6pt}})
\plot
        3.000   78.59   %% 1000
        3.301   84.67   %% 2000
        3.699   89.40   %% 5000
        4.000   91.32   %% 10000
        4.301   92.89   %% 20000
        4.699   94.46   %% 50000
        5.000   95.24   %% 100000
        5.301   95.81   %% 200000
        5.699   96.28   %% 500000
        6.000   96.65   %% 1000000
/
\setplotsymbol({\rule{.8pt}{.8pt}})
\plot
        3.000   95.17   %% 1000
        3.301   95.17   %% 2000
        3.699   95.56   %% 5000
        4.000   95.59   %% 10000
        4.301   95.77   %% 20000
        4.699   96.12   %% 50000
        5.000   96.42   %% 100000
        5.301   96.62   %% 200000
        5.699   96.78   %% 500000
        6.000   96.97   %% 1000000
/
\multiput {$\bullet$} at
        3.000   95.17   %% 1000
        3.301   95.17   %% 2000
        3.699   95.56   %% 5000
        4.000   95.59   %% 10000
        4.301   95.77   %% 20000
        4.699   96.12   %% 50000
        5.000   96.42   %% 100000
        5.301   96.62   %% 200000
        5.699   96.78   %% 500000
        6.000   96.91   %% 1000000
/
\plot
        3.000   62.18   %% 1000
        3.301   70.62   %% 2000
        3.699   77.12   %% 5000
        4.000   79.66   %% 10000
        4.301   81.54   %% 20000
        4.699   83.54   %% 50000
        5.000   84.38   %% 100000
        5.301   84.89   %% 200000
        5.699   85.30   %% 500000
        6.000   85.46   %% 1000000
/
\multiput {$\circ$} at
        3.000   62.18   %% 1000
        3.301   70.62   %% 2000
        3.699   77.12   %% 5000
        4.000   79.66   %% 10000
        4.301   81.54   %% 20000
        4.699   83.54   %% 50000
        5.000   84.38   %% 100000
        5.301   84.89   %% 200000
        5.699   85.30   %% 500000
        6.000   85.46   %% 1000000
/
\endpicture
\end{center}
\hrule
\caption{Learning curve for tagging the Penn Treebank. The training sets
  of variable sizes as well as test sets of 100,000 tokens were
  randomly chosen. Training and test sets were disjoint, the procedure
  was repeated 10 times and results were averaged.}
\label{fig:penn-learn}
\bigskip
\hrule
\begin{center}
{\bf Penn Treebank: Accuracy of reliable assignments}

\bigskip
\hspace*{0pt}
\beginpicture
\setcoordinatesystem units <23mm,10mm>
\setplotarea x from 0 to 4, y from 96 to 100
\axis bottom ticks withvalues 1 2 5 10 20 50 100 500 2000 10000 / 
                           at 0 .3 .7 1 1.3 1.7 2 2.7 3.3 4 / /
\axis bottom ticks at 2.3 3 3.7 / /
\axis left ticks numbered from 96 to 100 by 1 /
\linethickness=.4pt
\putbar breadth <0pt> from 0 97 to 4 97
\putbar breadth <0pt> from 0 98 to 4 98
\putbar breadth <0pt> from 0 99 to 4 99
\putbar breadth <0pt> from 0 100 to 4 100
\put {\rotate[l]{Accuracy}} [r] <-9mm,0mm> at 0 98
\put {threshold $\theta$} [lt] <7mm,-2.6mm> at 4 96
\put {\small 100} [t] <0mm,-8.5mm> at 0 96 
\put {\small 97.7} [t] <0mm,-8.5mm> at .3 96 
\put {\small 94.6} [t] <0mm,-8.5mm> at .7 96 
\put {\small 92.2} [t] <0mm,-8.5mm> at 1 96 
\put {\small 89.8} [t] <0mm,-8.5mm> at 1.3 96 
\put {\small 86.3} [t] <0mm,-8.5mm> at 1.7 96 
\put {\small 83.5} [t] <0mm,-8.5mm> at 2 96 
\put {\small 80.4} [t] <0mm,-8.5mm> at 2.3 96 
\put {\small 76.6} [t] <0mm,-8.5mm> at 2.7 96 
\put {\small 73.8} [t] <0mm,-8.5mm> at 3 96 
\put {\small 71.0} [t] <0mm,-8.5mm> at 3.3 96 
\put {\small 67.2} [t] <0mm,-8.5mm> at 3.7 96 
\put {\small 64.5} [t] <0mm,-8.5mm> at 4 96 
\put {\% cases reliable} [lt] <7mm,-8.5mm> at 4 96
\put {\small --} [t] <0mm,-13.5mm> at 0 96 
\put {\small 53.5} [t] <0mm,-13.5mm> at .3 96 
\put {\small 62.8} [t] <0mm,-13.5mm> at .7 96 
\put {\small 68.9} [t] <0mm,-13.5mm> at 1 96 
\put {\small 73.9} [t] <0mm,-13.5mm> at 1.3 96 
\put {\small 79.3} [t] <0mm,-13.5mm> at 1.7 96 
\put {\small 82.6} [t] <0mm,-13.5mm> at 2 96 
\put {\small 85.2} [t] <0mm,-13.5mm> at 2.3 96 
\put {\small 87.5} [t] <0mm,-13.5mm> at 2.7 96 
\put {\small 88.8} [t] <0mm,-13.5mm> at 3 96 
\put {\small 89.8} [t] <0mm,-13.5mm> at 3.3 96 
\put {\small 91.0} [t] <0mm,-13.5mm> at 3.7 96 
\put {\small 91.6} [t] <0mm,-13.5mm> at 4 96 
\put {acc. of complement} [lt] <7mm,-13mm> at 4 96
\put {\tabcolsep=0pt\begin{tabular}{llcr}
    \hbox to 0pt{\hspace*{4mm}$\diamond$\hss}%
    \rule[.5ex]{10mm}{1.6pt}\ \ \ & \multicolumn{3}{l}{Overall} \\
                             & \small min &\small=&\small 96.6\% \\
                             & \small max &\small=&\small 99.4\% \\
    \end{tabular}} [lb] <5mm,0mm> at 4 98.5
\setlinear
\setplotsymbol({\rule{1.6pt}{1.6pt}})
\plot
        0.00    96.63
        0.30    97.71
        0.70    98.61
        1.00    99.01
        1.30    99.25
        1.70    99.41
        2.00    99.46
        2.30    99.48
        2.70    99.49
        3.00    99.48
        3.30    99.46
        3.70    99.44
        4.00    99.42
/
\multiput {$\diamond$} at
        0.00    96.63
        0.30    97.71
        0.70    98.61
        1.00    99.01
        1.30    99.25
        1.70    99.41
        2.00    99.46
        2.30    99.48
        2.70    99.49
        3.00    99.48
        3.30    99.46
        3.70    99.44
        4.00    99.42
/
\endpicture
\medskip
\end{center}
\hrule
\caption{Tagging accuracy for the Penn Treebank when separating
  reliable and unreliable assignments. The curve shows accuracies for
  reliable assignments. The numbers at the bottom line
  indicate the percentage of reliable assignments and the accuracy of the complement set.}
\label{fig:penn-cond}
\end{figure*}

We exploit the fact that the tagger not only determines tags, but also
assigns probabilities. If there is an alternative that has a
probability ``close to'' that of the best assignment, this alternative
can be viewed as almost equally well suited.  The notion of ``close
to'' is expressed by the distance of probabilities, and this in turn
is expressed by the quotient of probabilities. So, the distance of the
probabilities of a best tag $t_{best}$ and an alternative tag
$t_{alt}$ is expressed by $p(t_{best}) / p(t_{alt})$, which is some
value greater or equal to 1 since the best tag assignment has the
highest probability.

Figure \ref{fig:negra-cond} shows the accuracy when separating
assignments with quotients larger and smaller than the threshold
(hence reliable and unreliable assignments). As expected, we find that
accuracies for reliable assignments are much higher than for unreliable
assignments. This distinction is, e.g., useful for annotation projects
during the cleaning process, or during pre-processing, so the tagger
can emit multiple tags if the best tag is classified as unreliable.

\subsection{Tagging the Penn Treebank}

We use the Wall Street Journal as contained in the Penn Treebank for
our experiments. The annotation consists of four parts: 1) a
context-free structure augmented with traces to mark movement and
discontinuous constituents, 2) phrasal categories that are annotated
as node labels, 3) a small set of grammatical functions that are
annotated as extensions to the node labels, and 4) part-of-speech tags
\cite{Marcus:ea:93}. This evaluation only uses the part-of-speech
annotation.

The Wall Street Journal part of the Penn Treebank
consists of approx.\ 50,000 sentences (1.2 million tokens).

Tagging accuracies for the Penn Treebank are shown in table
\ref{tab:pennacc}. 
Figure \ref{fig:penn-learn} shows the learning curve of the tagger,
i.e., the accuracy depending on the amount of training data. Training
length is the number of tokens used for training. Each training length
was tested ten times. Training and test sets were disjoint, results
are averaged. The training length is given on a logarithmic scale.
As for the NEGRA corpus, tagging accuracy is very high for known
tokens even with small amounts of training data.

We exploit the fact that the tagger not only determines tags, but also
assigns probabilities.  Figure \ref{fig:penn-cond} shows the accuracy
when separating assignments with quotients larger and smaller than the
threshold (hence reliable and unreliable assignments). Again, we find
that accuracies for reliable assignments are much higher than for
unreliable assignments.

\subsection{Summary of Part-of-Speech Tagging Results}

Average part-of-speech tagging accuracy is between 96\% and 97\%,
depending on language and tagset, which is at least on a par with
state-of-the-art results found in the literature, possibly better. For
the Penn Treebank, \cite{Ratnaparkhi:96} reports an accuracy of 96.6\%
using the Maximum Entropy approach, our much simpler and therefore
faster HMM approach delivers 96.7\%. This comparison needs to be 
re-examined, since we use a ten-fold crossvalidation and
averaging of results while Ratnaparkhi only makes one test run.

The accuracy for known tokens is significantly higher than for unknown
tokens. For the German newspaper data, results are 8.7\% better
when the word was seen before and therefore is in the lexicon, than
when it was not seen before (97.7\% vs.\ 89.0\%). Accuracy for known
tokens is high even with very small amounts of training data. As few
as 1000 tokens are sufficient to achieve 95\%--96\% accuracy for them.
It is important for the tagger to have seen a word at least once
during training.

Stochastic taggers assign probabilities to tags. We exploit the
probabilities to determine reliability of assignments. For a subset
that is determined during processing by the tagger we achieve accuracy rates
of over 99\%. The accuracy of the complement set is much lower.  This
information can, e.g., be exploited in an annotation project to give
an additional treatment to the unreliable assignments, or to pass
selected ambiguities to a subsequent processing step.

\section{Conclusion}
\label{sec:conclusion}

We have shown that a tagger based on Markov models yields
state-of-the-art results, despite contrary claims found in the
literature. For example, the Markov model tagger used in the
comparison of \cite{Halteren:ea:98} yielded worse results than all
other taggers. In our opinion, a reason for the wrong claim is that
the basic algorithms leave several decisions to the implementor. The
rather large amount of freedom was not handled in detail in previous
publications: handling of start- and end-of-sequence, the exact
smoothing technique, how to determine the weights for context
probabilities, details on handling unknown words, and how to determine
the weights for unknown words.  Note that the decisions we made yield
good results for both the German and the English Corpus. They do so
for several other corpora as well.  The architecture remains
applicable to a large variety of languages.

According to current tagger comparisons
\cite{Halteren:ea:98,Zavrel:Daelemans:99}, and according to a
comparsion of the results presented here with those in
\cite{Ratnaparkhi:96}, the Maximum Entropy framework seems to be the
only other approach yielding comparable results to the one presented
here. It is a very interesting future research topic to determine the
advantages of either of these approaches, to find the reason for their
high accuracies, and to find a good combination of both.

\tnt\ is freely available to universities and related organizations
for research purposes (see {\tt
  http://www.coli.uni-sb.de/\char126thorsten/tnt}).

\section*{Acknowledgements}

Many thanks go to Hans Uszkoreit for his support during the
development of \tnt. Most of the work on \tnt\ was carried out while
the author received a grant of the Deutsche Forschungsgemeinschaft in
the Graduiertenkolleg Kognitionswissenschaft Saarbr{\"u}cken. Large
annotated corpora are the pre-requisite for developing and testing
part-of-speech taggers, and they enable the generation of high-quality
language models.  Therefore, I would like to thank all the people who
took the effort to annotate the Penn Treebank, the Susanne Corpus, the
Stutt\-garter Referenzkorpus, the NEGRA Corpus, the Verbmobil Corpora,
and several others.  And, last but not least, I would like to thank
the users of \tnt\ who provided me with bug reports and valuable
suggestions for improvements.


\begin{thebibliography}{}

\bibitem[\protect\citename{Brants \bgroup et al.\egroup }1999]{Brants:ea:99}
Thorsten Brants, Wojciech Skut, and Hans Uszkoreit.
\newblock 1999.
\newblock Syntactic annotation of a {G}erman newspaper corpus.
\newblock In {\em Proceedings of the {ATALA} Treebank Workshop}, pages 69--76,
  Paris, France.

\bibitem[\protect\citename{Brill}1993]{Brill:93diss}
Eric Brill.
\newblock 1993.
\newblock {\em A Corpus-Based Approach to Language Learning}.
\newblock Ph.D. Dissertation, Department of Computer and Information Science,
  University of Pennsylvania.

\bibitem[\protect\citename{Charniak \bgroup et al.\egroup
  }1993]{Charniak:ea:93}
Eugene Charniak, Curtis Hendrickson, Neil Jacobson, and Mike Perkowitz.
\newblock 1993.
\newblock Equations for part-of-speech tagging.
\newblock In {\em Proceedings of the Eleventh National Conference on Artificial
  Intelligence}, pages 784--789, Menlo Park: AAAI Press/MIT Press.

\bibitem[\protect\citename{Cutting \bgroup et al.\egroup }1992]{Cutting:ea:92}
Doug Cutting, Julian Kupiec, Jan Pedersen, and Penelope Sibun.
\newblock 1992.
\newblock A practical part-of-speech tagger.
\newblock In {\em Proceedings of the 3rd Conference on Applied Natural Language
  Processing ({ACL})}, pages 133--140.

\bibitem[\protect\citename{Daelemans \bgroup et al.\egroup
  }1996]{Daelemans:ea:96}
Walter Daelemans, Jakub Zavrel, Peter Berck, and Steven Gillis.
\newblock 1996.
\newblock Mbt: A memory-based part of speech tagger-generator.
\newblock In {\em Proceedings of the Workshop on Very Large Corpora},
  Copenhagen, Denmark.

\bibitem[\protect\citename{Marcus \bgroup et al.\egroup }1993]{Marcus:ea:93}
Mitchell Marcus, Beatrice Santorini, and Mary~Ann Marcinkiewicz.
\newblock 1993.
\newblock Building a large annotated corpus of {E}nglish: {T}he {P}enn
  {T}reebank.
\newblock {\em Computational Linguistics}, 19(2):313--330.

\bibitem[\protect\citename{Rabiner}1989]{Rabiner:89}
Lawrence~R. Rabiner.
\newblock 1989.
\newblock A tutorial on {H}idden {M}arkov {M}odels and selected applications in
  speech recognition.
\newblock In {\em Proceedings of the {IEEE}}, volume 77(2), pages 257--285.

\bibitem[\protect\citename{Ratnaparkhi}1996]{Ratnaparkhi:96}
Adwait Ratnaparkhi.
\newblock 1996.
\newblock A maximum entropy model for part-of-speech tagging.
\newblock In {\em Proceedings of the Conference on Empirical Methods in Natural
  Language Processing {EMNLP}-96}, Philadelphia, PA.

\bibitem[\protect\citename{Samuelsson}1993]{Samuelsson:93}
Christer Samuelsson.
\newblock 1993.
\newblock Morphological tagging based entirely on {B}ayesian inference.
\newblock In {\em 9th Nordic Conference on Computational Linguistics
  {NODALIDA}-93}, Stockholm University, Stockholm, Sweden.

\bibitem[\protect\citename{Schmid}1995]{Schmid:95}
Helmut Schmid.
\newblock 1995.
\newblock Improvements in part-of-speech tagging with an application to
  {G}erman.
\newblock In Helmut Feldweg and Erhard Hinrichts, editors, {\em {Lexikon und
  Text}}. Niemeyer, T{\"u}bingen.

\bibitem[\protect\citename{Skut \bgroup et al.\egroup }1997]{Skut:ea:97a}
Wojciech Skut, Brigitte Krenn, Thorsten Brants, and Hans Uszkoreit.
\newblock 1997.
\newblock An annotation scheme for free word order languages.
\newblock In {\em Proceedings of the Fifth Conference on Applied Natural
  Language Processing {ANLP}-97}, Washington, DC.

\bibitem[\protect\citename{van Halteren \bgroup et al.\egroup
  }1998]{Halteren:ea:98}
Hans van Halteren, Jakub Zavrel, and Walter Daelemans.
\newblock 1998.
\newblock Improving data driven wordclass tagging by system combination.
\newblock In {\em Proceedings of the International Conference on Computational
  Linguistics {COLING}-98}, pages 491--497, Montreal, Canada.

\bibitem[\protect\citename{Volk and Schneider}1998]{Volk:Schneider:98}
Martin Volk and Gerold Schneider.
\newblock 1998.
\newblock Comparing a statistical and a rule-based tagger for german.
\newblock In {\em Proceedings of {KONVENS}-98}, pages 125--137, Bonn.

\bibitem[\protect\citename{Zavrel and Daelemans}1999]{Zavrel:Daelemans:99}
Jakub Zavrel and Walter Daelemans.
\newblock 1999.
\newblock Evaluatie van part-of-speech taggers voor het corpus gesproken
  nederlands.
\newblock {CGN} technical report, Katholieke Universiteit Brabant, Tilburg.

\end{thebibliography}
\end{document}